\documentclass[10pt,twocolumn,letterpaper]{article}

\usepackage{cvpr}
\usepackage{times}
\usepackage{epsfig}
\usepackage{graphicx}
\usepackage{amsmath}
\usepackage{amssymb}

\usepackage{amsthm}
\usepackage{mathrsfs} 

\usepackage[linesnumbered,ruled]{algorithm2e}
\usepackage[pagebackref=true,breaklinks=true,letterpaper=true,colorlinks,bookmarks=false]{hyperref}

\usepackage[utf8]{inputenc} 
\usepackage[T1]{fontenc}    
\usepackage{hyperref}       
\usepackage{url}            
\usepackage{booktabs}       
\usepackage{amsfonts}       
\usepackage{nicefrac}       
\usepackage{microtype}      

\DeclareMathOperator*{\argmin}{argmin}
\DeclareMathOperator*{\argmax}{argmax}

\usepackage[pagebackref=true,breaklinks=true,letterpaper=true,colorlinks,bookmarks=false]{hyperref}

\cvprfinalcopy 

\def\cvprPaperID{****} 

\ifcvprfinal\fi
\begin{document}

\title{Sliced Wasserstein Distance for Learning Gaussian Mixture Models}

\author{Soheil Kolouri\\
HRL Laboratories, LLC\\
{\tt\small skolouri@hrl.com}
\and
Gustavo K. Rohde\\
University of Virginia\\
{\tt\small gustavo@virginia.edu}
\and
Heiko Hoffmann\\
HRL Laboratories, LLC\\
{\tt\small hhoffmann@hrl.com}}

\maketitle

\begin{abstract}
 Gaussian mixture models (GMM) are powerful parametric tools with many applications in machine learning and computer vision. Expectation maximization (EM) is the most popular algorithm for estimating the GMM parameters. However, EM  guarantees only convergence to a stationary point of the log-likelihood function, which could be arbitrarily worse than the optimal solution. Inspired by the relationship between the negative log-likelihood function and the Kullback-Leibler (KL) divergence, we propose an alternative formulation for estimating the GMM parameters using the sliced Wasserstein distance, which gives rise to a new algorithm. Specifically, we propose minimizing the sliced-Wasserstein distance between the mixture model and the data distribution with respect to the GMM parameters. In contrast to the KL-divergence, the energy landscape for the sliced-Wasserstein distance is more well-behaved and therefore more suitable for a stochastic gradient descent scheme to obtain the optimal GMM parameters. We show that our formulation results in parameter estimates that are more robust to random initializations and demonstrate that it can estimate high-dimensional data distributions more faithfully than the EM algorithm.
\end{abstract}

\section{Introduction}
\label{sec:introduction}

Finite Gaussian Mixture Models (GMMs), also called Mixture of Gaussians (MoG), are powerful, parametric, and probabilistic tools that are widely used as flexible models for multivariate density estimation in various applications concerning machine learning, computer vision, and signal/image analysis. GMMs have been utilized for: image representation \cite{beecks2011modeling,goldberger2003efficient} to generate feature signatures, point set registration \cite{jian2011robust}, adaptive contrast enhancement \cite{celik2012automatic}, inverse problems including super-resolution and deblurring \cite{guerrero2008image,yu2012solving}, time series classification \cite{campbell2006support}, texture segmentation \cite{permuter2006study}, and robotic visuomotor transformations \cite{hoffmann05bc} among many others.  

As a special case of general latent variable models, finite GMM parameters could serve as a concise embedding \cite{murphy2012machine}, which provide a compressed representation of the data. Moreover, GMMs could be used to approximate any density defined on $\mathbb{R}^d$ with a large enough number of mixture components. To fit a finite GMM to the observed data, one is required to answer the following questions: 1) how to estimate the number of mixture components needed to represent the data, and 2) how to estimate the parameters of the mixture components. Several techniques have been introduced to provide an answer for the first question \cite{mclachlan2014number}. The focus of this paper in on the latter question. 

The existing methods to estimate the GMM parameters are based on minimizing the negative log-likelihood (NLL) of the data with respect to the parameters \cite{tipping+99b}. The Expectation Maximization (EM) algorithm \cite{dempster1977maximum} is the prominent way of minimizing the NLL (though, see, e.g., as an alternative \cite{moeller+04,hoffmann05ulva}). While EM remains the most popular method for estimating GMMs, it only guarantees convergence to a stationary point of the likelihood function. On the other hand, various studies have shown that the likelihood function has bad local maxima that can have arbitrarily worse log-likelihood values compared to any of the global maxima \cite{hoffmann05ulva,jin2016local,amendola2015maximum}. More importantly, Jin et al. \cite{jian2011robust} proved that with random initialization, the EM algorithm will converge to a bad critical point with high probability. This issue turns the EM algorithm sensitive to the choice of initial parameters. 

In the limit (i.e. having infinite i.i.d samples), minimizing the NLL function is equivalent to minimizing the Kullback-Leibler divergence between the data distribution and the GMM, with respect to the GMM parameters. Here, we propose, alternatively, to minimize the p-Wasserstein distance, and more specifically the sliced p-Wasserstein distance \cite{kolouri2017optimal}, between the data distribution and the GMM. The Wasserstein distance and its variations have attracted a lot of attention from the Machine Learning (ML) and signal processing communities lately \cite{kolouri2017optimal,arjovsky2017wasserstein,frogner2015learning}. It has been shown that optimizing with respect to the Wasserstein loss has various practical benefits over the KL-divergence loss \cite{peyre2012wasserstein,frogner2015learning,montavon2016wasserstein,arjovsky2017wasserstein,gulrajani2017improved}. Importantly, unlike the KL-divergence and its related dissimilarity measures (e.g. Jensen-Shannon divergence), the Wasserstein distance can provide a meaningful notion of closeness (i.e. distance) for distributions supported on non-overlapping low dimensional manifolds. This motivates our proposed formulation for estimating GMMs. 

To overcome the computational burden of the Wasserstein minimization for high-dimensional distributions, we propose to use the sliced Wasserstein distance \cite{bonneel2015sliced,kolouri2016sliced,kolouri2017optimal}. Our method slices the high-dimensional data distribution via random projections and minimizes the Wasserstein distance between the projected one-dimensional distributions with respect to the GMM parameters. We note that the idea of characterizing a high-dimensional distribution via its random projections has been studied in various work before \cite{vempala2005random,kalai2012disentangling}. The work in \cite{kalai2012disentangling}, for instance, minimizes the $L_1$ norm between the slices of the data distribution and the GMM with respect to the parameters. This method, however, suffers from the same shortcomings as the KL-divergence based methods.

The p-Wasserstein distances and more generally the optimal mass transportation problem  have recently gained plenty of attention from the computer vision and machine learning communities \cite{kolouri2017optimal,thorpe2017transportation,park2017cumulative,kolouri2015transport,xiao2017wasserstein,rolet2016fast,arjovsky2017wasserstein}.  We note that the p-Wasserstein distances have also been used in regard to GMMs, however, as a distance metric to compare various GMM models \cite{chen2017optimal,li2013novel,qian2017deep}. Our proposed method, on the other hand, is an alternative framework for fitting a GMM to data via sliced p-Wasserstein distances. 

In what follows, we first formulate the p-Wasserstein distance, the Radon transform, and the Sliced p-Wasserstein distance in Section \ref{sec:prelim}. In Section \ref{sec:SWM}, we reiterate the connection between the K-means problem and the Wasserstein means problem \cite{ho2017multilevel}, extend it to GMMs, and formulate the Sliced Wasserstein means problem. Our numerical experiments are presented in Section \ref{sec:experiments}. Finally, we conclude our paper in Section \ref{sec:conclusion}. 


\section{Preliminary}
\label{sec:prelim}
\subsection{p-Wasserstein distance:}
In this section we review the preliminary concepts and formulations needed to develop our framework. Let $P_p(\Omega)$ be the set of Borel probability measures with finite $p$'th moment defined on a given metric space $(\Omega,d)$,  and let $\rho\in P_p(X)$ and $\nu\in P_p(Y)$ be probability measures defined on $X,Y\subseteq \Omega$ with corresponding probability density functions $I_x$ and $I_y$, $d\rho(x)=I_x(x)dx$ and $d\nu(y)=I_y(y)dy$. The $p$-Wasserstein distance for $p\in[1,\infty)$ between $\rho$ and $\nu$ is defined as the optimal mass transportation  (OMT) problem \cite{villani2008optimal}  with cost function $c(x,y)=d^p(x,y)$, such that:
\begin{eqnarray}
W_p(\rho,\nu)=\left(\operatorname*{inf}_{\gamma\in \Gamma(\rho,\nu)} \int_{X\times Y} d^p(x,y)d \gamma(x,y) \right)^{\frac{1}{p}},
\label{eq:kantorovich}
\end{eqnarray}
where $ \Gamma(\rho,\nu)$ is the set of all transportation plans, $\gamma\in\Gamma(\rho,\nu)$, and satisfy the following:
\begin{eqnarray*}
\begin{array}{lr}
\gamma(A \times Y)= \rho(A) & \text{for any Borel subset } A\subseteq X\\
\gamma(X \times B)= \nu(B) & \text{for any Borel subset } B\subseteq Y
\end{array}.
\end{eqnarray*}
Due to Brenier's theorem \cite{brenier1991polar}, for  absolutely continuous probability measures $\rho$ and $\nu$   (with respect to Lebesgue measure) the $p$-Wasserstein distance can be  equivalently obtained from, 
\begin{eqnarray}
W_p(\rho,\nu)=(\operatorname{inf}_{f\in MP(\rho,\nu)} \int_{X} d^p(f(x),x) d\rho(x))^{\frac{1}{p}}
\end{eqnarray}
where, $MP(\rho,\nu)=\{ f:X\rightarrow Y ~|~ f_\#\rho=\nu\}$ and $f_\#\rho$ represents the pushforward of measure $\rho$,
\begin{equation*}
\int_{f^{-1}(A)} d\rho(x)= \int_{A}d\nu(y) ~\text{for any Borel subset } A\subseteq Y.
\end{equation*}

When a transport map exists, the transport plan and the transport map are related via, $\gamma=(\mathrm{Id}\times f)_\#\rho$.
Note that in most engineering and computer science applications $\Omega$ is a compact subset of $\mathbb{R}^d$ and  $d(x,y)=|x-y|$ is the Euclidean distance. By abuse of notation we will use $W_p(\rho, \nu)$ and $W_p(I_x,I_y)$ interchangeably throughout the manuscript. For a more detailed explanation of the Wasserstein distances and the optimal mass transport problem, we refer the reader to the recent review article by Kolouri et al. \cite{kolouri2017optimal} and the references there in.

{\bf One-dimensional distributions:}  The case of one-dimensional continuous probability measures is specifically interesting as the p-Wasserstein distance has a closed form solution. More precisely, for one-dimensional probability measures there exists a unique monotonically increasing transport map that pushes one measure into another. Let $J_x(x)=\rho((-\infty,x])=\int_{-\infty}^{x} I_x(\tau)d\tau$ be the cumulative distribution function (CDF) for $I_x$ and define $J_y$ to be the CDF of $I_y$. The transport map is then uniquely defined as, $f(x)= J_y^{-1}(J_x(x))$ and consequently the $p$-Wasserstein distance is calculated as: 
\begin{eqnarray}
W_p(\rho,\nu)&=&\left( \int_{X} d^p(J_y^{-1}(J_x(x)),x) d\rho(x) \right)^{\frac{1}{p}}\nonumber\\
&=& \left(\int_{0}^1 d^p(J_y^{-1}(z),J_x^{-1}(z))dz \right)^{\frac{1}{p}}
\label{eq:1dwp}
\end{eqnarray}
where in the second line we used the change of variable $J_x(x)=z$. The closed form solution of the p-Wasserstein is an attractive property that gives rise to the Sliced-Wasserstein (SW) distances. Next we review the Radon transform, which enables the definition the Sliced $p$-Wasserstein distance.

\subsection{ Radon transform}  

 The $d$-dimensional Radon transform, $\mathcal{R}$, maps a function  $I\in L^1(\mathbb{R}^d)$ where $L^1(\mathbb{R}^d):=\{ I:\mathbb{R}^d \rightarrow \mathbb{R} | \int_{\mathbb{R}^d} |I(x)|dx \leq \infty\}$ to the set of its integrals over the hyperplanes of $\mathbb{R}^d$ and is defined as, 
\begin{eqnarray}
\mathcal{R} I(t,\theta):=\int_{\mathbb{R}^d} I(x)\delta(t-x\cdot\theta)dx
\label{eq:radon}
\end{eqnarray}
For all  $\theta\in \mathbb{S}^{d-1}$ where $\mathbb{S}^{d-1}$ is the unit sphere in $\mathbb{R}^{d}$. Note that $\mathcal{R}: L^1(\mathbb{R}^d)\rightarrow L^1(\mathbb{R}\times \mathbb{S}^{d-1})$. For the sake of completeness, we note that the Radon transform is an invertible, linear transform and we denote its inverse as $\mathcal{R}^{-1}$, which is also known as the filtered back projection algorithm and is defined as:
\begin{eqnarray}
I(x)&=&\mathcal{R}^{-1}(\mathcal{R}I(t,\theta))
\nonumber\\&=& \int_{\mathbb{S}^{d-1}} (\mathcal{R}I(.,\theta)*h(.))\circ (x\cdot\theta)d\theta
\label{eq:backproj}
\end{eqnarray}
where $h(.)$ is a one-dimensional filter with corresponding Fourier transform $\mathcal{F}h(\omega)=c|\omega|^{d-1}$ (it appears due to the Fourier slice theorem, see \cite{natterer1986mathematics} for more details) and `$*$' denotes convolution. Radon transform and its inverse are extensively used in Computerized Axial Tomography (CAT) scans in the field of medical imaging, where X-ray measurements integrate the tissue-absorption levels along 2D hyper-planes to provide a tomographic image of the internal organs. Note that in practice acquiring infinite number of projections is not feasible therefore the integration in Equation \eqref{eq:backproj} is replaced with a finite summation over projection angles. A formal measure theoretic definition of Radon transform for probability measures could be find in \cite{bonneel2015sliced}. 

{\bf Radon transform of empirical PDFs:} The Radon transform of $I_x$ simply follows Equation \eqref{eq:radon}. However, in most machine learning applications we do not have access to the distribution $I_x$ but to its samples, $x_n$. Kernel density estimation could be used in such scenarios to approximate $I_x$ from its samples,
\begin{eqnarray*}
I_x(x)\approx \frac{1}{N_\rho}\sum_{n=1}^{N_\rho} \phi(x-x_n)
\end{eqnarray*} 
where $\phi:\mathbb{R}^d\rightarrow \mathbb{R}^+$ is a density kernel where $\int_{\mathbb{R}^d} \phi(x)dx=1$ (e.g. Gaussian kernel). The Radon transform of  $I_x$ can then be approximated from its samples via:
\begin{eqnarray}
\mathcal{R}I_x(t,\theta)\approx \frac{1}{N_\rho}\sum_{n=1}^{N_\rho} \mathcal{R}\phi(t-x_n\cdot\theta,\theta)
\label{eq:emplinear}
\end{eqnarray} 

 Note that certain density kernels have analytic Radon transformation. For instance when $\phi(x)=\delta(x)$ the Radon transform $\mathcal{R}\phi(t,\theta)=\delta(t)$. 
 
 {\bf Radon transform of multivariate Gaussians:}  Let $\phi(x)=\mathcal{N}_d(\mu,\Sigma)$ be a d-dimensional multivariate Gaussian distribution with mean $\mu\in \mathbb{R}^d$ and covariance $\Sigma\in \mathbb{R}^{d\times d}$. A slice/projection of the Radon transform of $\phi$ is then a one-dimensional normal distribution $\mathcal{R}\phi(\cdot,\theta)=\mathcal{N}_1(\theta\cdot x,\theta^T\Sigma\theta)$. Given the linearity of the Radon transform, this indicates that a slice of a d-dimensional GMM is a one-dimensional GMM with component means $\theta\cdot\mu_i$ and variance $\theta^T\Sigma_i\theta$.

\subsection{ Sliced $p$-Wasserstein Distance}  

The idea behind the sliced $p$-Wasserstein distance is to first obtain a family of marginal distributions (i.e. one-dimensional distributions) for a higher-dimensional probability distribution through linear projections (via Radon transform), and then calculate the distance between two input distributions as a functional on the $p$-Wasserstein distance of their marginal distributions. In this sense, the distance is obtained by solving several one-dimensional optimal transport problems, which have closed-form solutions. More precisely, the Sliced Wasserstein distance between $I_x$ and $I_y$ is defined as, 
 \begin{eqnarray}
 SW_p(I_x,I_y)=(\int_{\mathbb{S}^{d-1}} W^p_p(\mathcal{R} I_x(.,\theta),\mathcal{R} I_y(.,\theta)) d\theta)^{\frac{1}{p}}
 \label{eq:SW}
 \end{eqnarray}
 The Sliced $p$-Wasserstein distance as defined above is symmetric, and it satisfies sub-additivity and coincidence axioms, and hence it is a true metric \cite{kolouri2016sliced}. 
 
 The sliced $p$-Wasserstein distance is especially useful when one only has access to samples of a high-dimensional PDFs and kernel density estimation is required to estimate $I$. One dimensional kernel density estimation of PDF slices is a much simpler task compared to direct estimation of $I$ from its samples. The catch, however, is that as the dimensionality grows one requires  larger number of projections to estimate $I$ from $\mathcal{R}I(., \theta)$. In short, if a reasonably smooth two-dimensional distribution can be approximated by its $L$ projections (up to an acceptable reconstruction error, $\epsilon$), then one would require $\mathcal{O}(L^{d-1})$ number of projections to approximate a similarly smooth d-dimensional distribution (for $d\geq2$). In later sections we show that the projections could be randomized in a stochastic Gradient descent fashion for learning Gaussian mixture models. 


\section{Sliced Wasserstein Means and Gaussian Mixture Models}
\label{sec:SWM}

Here we first reiterate the connection between the K-means clustering algorithm and the Wasserstein means problem, and then extend this connection to GMMs and state the need for the sliced Wasserstein distance. Let $y_n\in \mathbb{R}^d$ for $n=1,...,N$ be $N$  samples and $Y=[y_1,...,y_N]\in \mathbb{R}^{d\times N}$. The K-means clustering algorithm seeks the best $K$ points, $x_k\in \mathbb{R}^d$ for $k=1,...,K$ and $X=[x_1,...,x_K]\in \mathbb{R}^{d\times K}$, that represent $Y$. Formally, 
\begin{eqnarray}
&&\operatorname{inf}_{C,X} \frac{1}{N} \|Y-XC^T\|^2\nonumber\\ 
&&~s.t.~~ {C1_K=1_N,c_{i,j}\in \{0,1\}}
\end{eqnarray}
where $C\in\mathbb{R}^{N\times K}$ contains the one-hot labels of $Y$. 

Let $I_y=\frac{1}{N}\sum_{n=1}^N \phi (y-y_n)$ be the empirical distribution of $Y$, where $\phi$ is a kernel density estimator (e.g. radial basis function kernel or the Dirac delta function in its limit). Then, the K-means problem can be alternatively solved by minimizing a statistical distance/divergence between $I_y$ and $I_x=\frac{1}{K}\sum_{k=1}^K \phi(x-x_k)$. A common choice for such distance/divergence is the Kullback-Leibler divergence (KL-divergence) \cite{banerjee2005clustering,chaudhuri2008finding}. Alternatively, the p-Wasserstein distance could be used to estimate the parameters of $I_x$, 
\begin{eqnarray}
\operatorname{inf}_{I_x} W_p^p(I_x,I_y)
\label{eq:WM}
\end{eqnarray}

We discuss the benefits of the p-Wasserstein distance over the KL-divergence in the next sub-section.  Above minimization is known as the {\it Wasserstein means problem} and is closely related to the Wasserstein Barycenter problem \cite{agueh2011barycenters,rabin2012wasserstein,cuturi2014fast,ho2017multilevel}. The main difference being in that in these works the goal is to find a measure  $\nu*$ such that $\nu* = \arg\inf_\nu \sum_k W_p^p(\nu_k,\nu)$, where $\nu_k$ are sets of given low dimensional distributions (2 or 3D images or point clouds). The strategy in \cite{agueh2011barycenters,rabin2012wasserstein,cuturi2014fast} could also be extended into a clustering problem, though the two formulations are still significantly different given the inputs into the wasserstein distance being very different. Note also that K-means is equivalent to a variational EM approximation of a GMM with isotropic Gaussians \cite{lucke2017k}, therefore, a natural extension of the {\it Wasserstein means problem} could be formulated to fit a general GMM to $I_y$. To do so, we let distribution $I_x$ to be the parametric GMM as follows: 
{\small\begin{equation*}
I_x(x)= \sum_k  \frac{\alpha_k}{(2\pi)^{\frac{d}{2}} \sqrt{det(\Sigma_k)}}exp(-\frac{1}{2}(x-\mu_k)^T\Sigma_k^{-1}(x-\mu_k)) 
\end{equation*}}
where $\sum_k\alpha_k=1$ and Equation \eqref{eq:WM} is solved to find $\mu_k$s, $\Sigma_k$s, and $\alpha_k$s. Next we describe the benefits of using the Wasserstein distance in Equation \eqref{eq:WM} to fit a general GMM to the observed data compared to the common log-likelihood maximization schemes.

\begin{figure}[t]
\centering
\includegraphics[width=\linewidth]{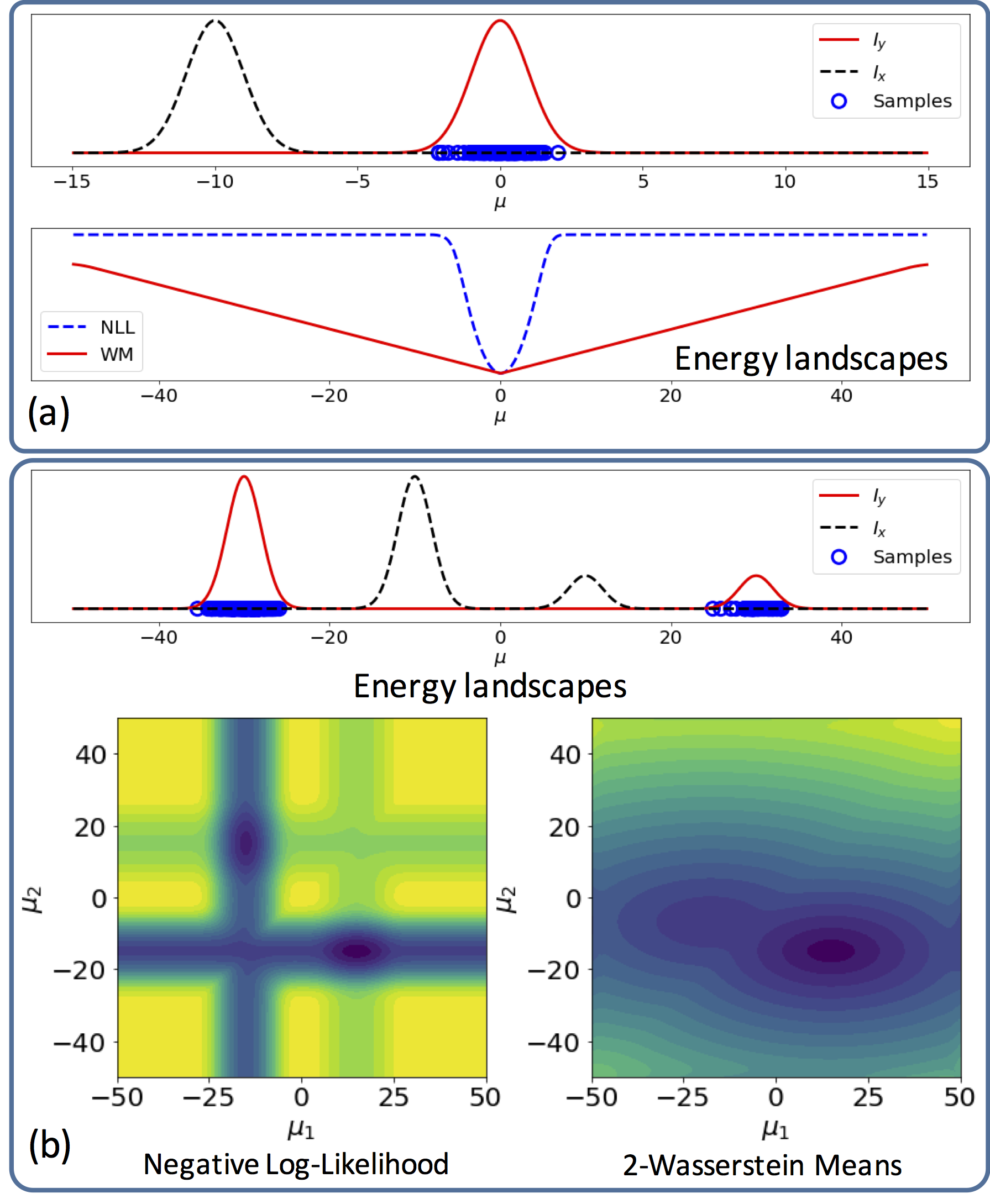}
\caption{The corresponding energy landscapes for the negative log-likelihood and the Wasserstein Means problem for scenario 1 (a) and scenario 2 (b). The energy landscapes are scaled and shifted for visualization purposes.}
\label{fig:energy}
\end{figure}

\subsection{Wasserstein Means vs. Maximum Log-Likelihood} 

General GMMs are often fitted to the observed data points, $y_n$s, via maximizing the log-likelihood of samples with respect to $I_x$. Formally, this is written as:
\begin{eqnarray}
\max_{\mu_k,\Sigma_k,\alpha_k} \frac{1}{N}\sum_{n=1}^N log(I_x(y_n))
\label{eq:lle}
\end{eqnarray}
It is straightforward to show that in the limit and as the number of samples grows to infinity, $N\rightarrow\infty$, the maximum log-likelihood becomes equivalent to minimizing the KL-divergence between $I_x$ and $I_y$ (See supplementary material for a proof):
\begin{eqnarray*}
\lim_{N\rightarrow \infty}\max_{\mu_k,\Sigma_k,\alpha_k} \frac{1}{N}\sum_{n=1}^N log(I_x(y_n))= \min_{\mu_k,\Sigma_k,\alpha_k}  KL(I_x,I_y)\nonumber\\
\end{eqnarray*}

The p-Wasserstein distance has been shown to have certain benefits over the commonly used KL-divergence and its related distances/divergences (i.e., other examples of Bregman divergences including the Jensen-Shannon (JS) distance and Itakura-Saito distance) \cite{arjovsky2017wasserstein}. For a general GMM, the model $I_x$ is continuous and smooth (i.e. infinitely differentiable) in its parameters and is locally Lipschitz; therefore, $W_p(I_x,I_y)$ is continuous and differentiable everywhere, while this is not true for the KL-divergence. In addition,  in scenarios where the distributions are supported by low dimensional manifolds, KL-divergence and other Bregman divergences may be difficult cost functions to optimize given their limited capture range. This limitation is due to their {\it `Eulerian'} nature, in the sense that the distributions are compared at fixed spatial coordinates (i.e., bin-to-bin comparison in discrete measures) as opposed to the p-Wasserstein distance, which is {\it `Lagrangian'}, and morphs one distribution to match another by finding correspondences in the domain of these distributions (i.e., Wasserstein distances perform cross-bin comparisons).

To get a practical sense of the benefits of the Wasserstein means problem over the maximum log-likelihood estimation, we study two simple scenarios. In the first scenario, we generate $N$ one-dimensional samples, $y_n$, from a normal distribution $\mathcal{N}(0,\sigma)$ where we assume known $\sigma$ and visualize the {\it negative log-likelihood (NLL)} and the {\it Wasserstein means (WM) problem} as a function of $\mu$. Figure \ref{fig:energy} (a) shows the first scenario and the corresponding energy landscapes for the negative log-likelihood and the Wasserstein means problem. It can be seen that while the global optimum is the same for both problems, the Wasserstein means landscape is less sensitive to the initial point, hence a gradient descent approach would easily converge to the optimal point regardless of the starting point. In the second scenario, we generated $N$ samples, $y_n$, from a mixture of two one-dimensional Gaussian distributions. Next, we assumed that the mixture coefficients $\alpha_k$s and the standard deviations $\sigma_k$s, for $k\in\{0,1\}$, are known and visualized the corresponding energy landscapes for NLL and WM as a function of $\mu_k$s (See Figure \ref{fig:energy} (b)). It can be clearly seen that although the global optimum of both problems is the same, but the energy landscape of the Wasserstein means problem does not suffer from local minima and is much smoother.

The Wasserstein means problem, however, suffers from the fact that the $W_2^2(.,.)$ is computationally expensive to calculate for high-dimensional $I_x$ and $I_y$. This is true even using very efficient OMT solvers, including the ones introduced by Cuturi \cite{cuturi2013sinkhorn}, Solomon et al. \cite{solomon2015convolutional}, and Levy \cite{levy2015}.

\begin{figure}[t]
\centering
\includegraphics[width=\linewidth]{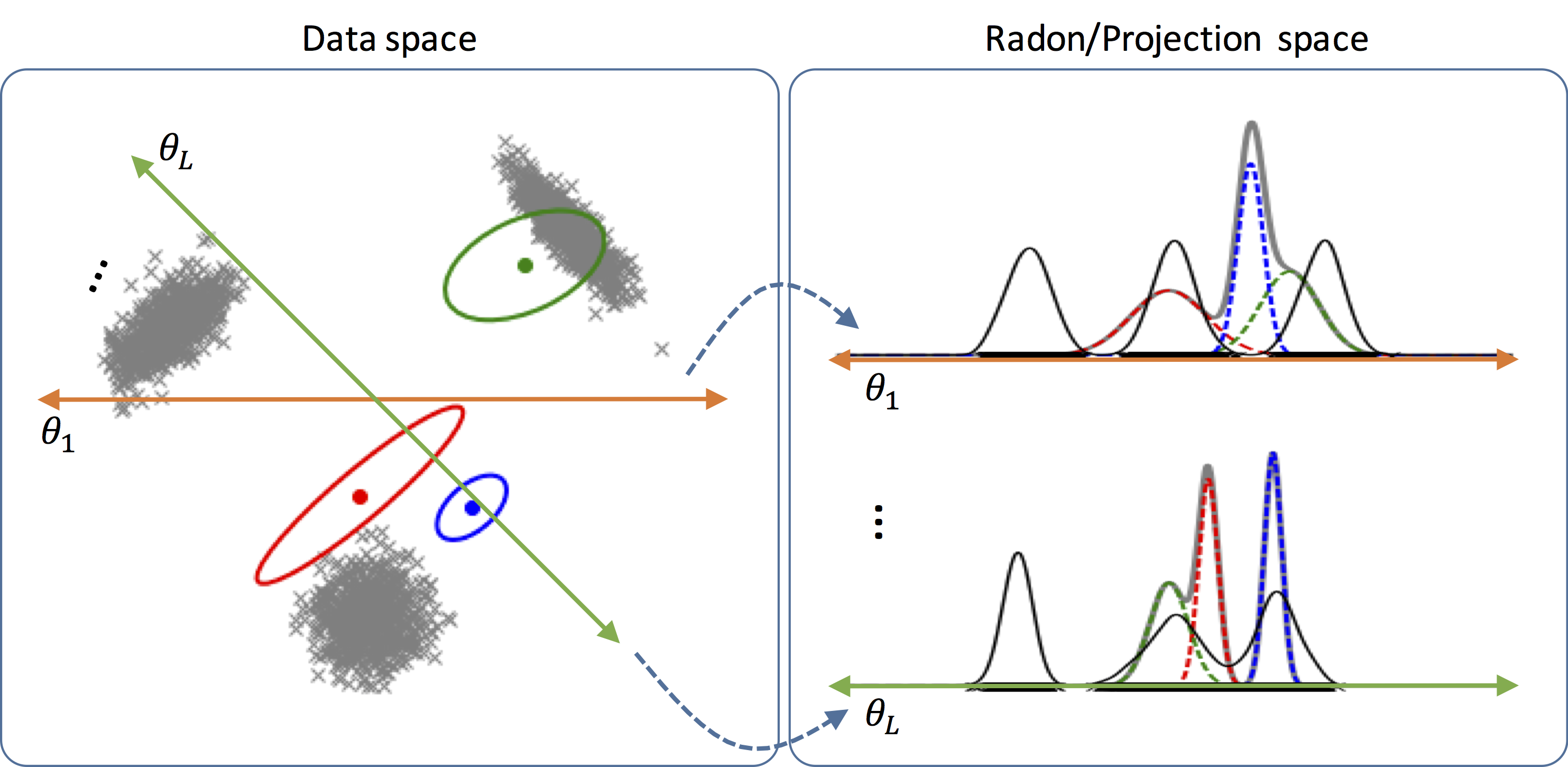}
\caption{Illustration of the high-level approach for the Sliced-Wasserstein Means of GMMs. }
\label{fig:hlidea}
\end{figure}

\subsection {Sliced Wasserstein Means} We propose to use an approximation of the p-Wasserstein distance between $I_x$ and $I_y$ using the SW distance. Substituting the Wasserstein distance in Equation \eqref{eq:WM} with the SW distance leads to the {\it Sliced p-Wasserstein Means (SWM)}  problem, 
{\small\begin{eqnarray*}
\operatorname*{inf}_{\mu_k,\Sigma_k,\alpha_k} SW^p_p(I_x,I_y)&=& \int_{\mathbb{S}^{d-1}} W_p^p(\mathcal{R}I_x(.,\theta),\mathcal{R} I_y(.,\theta)) d\theta,
\end{eqnarray*}}
which can be written as:
\begin{eqnarray}
\operatorname*{inf}_{\mu_k,\Sigma_k,\alpha_k}  \int_{\mathbb{S}^{d-1}}\operatorname{inf}_{f(.,\theta)} \int_\mathbb{R} |f(t,\theta)-t|^p \mathcal{R}I_x(t,\theta)dtd\theta \nonumber\\
\label{eq:SWM}
\end{eqnarray}
where for a fixed $\theta$, $f(.,\theta)$ is the optimal transport map between $\mathcal{R}I_x(.,\theta)$ and $\mathcal{R}I_y(.,\theta)$, and satisfies $\frac{\partial f(t,\theta) }{\partial t}\mathcal{R}I_y(f(t,\theta),\theta)=\mathcal{R}I_x(t,\theta)$. Note that, since $I_x$ is an absolutely continuous PDF, an optimal transport map will exist even when $I_y$ is not an absolutely continuous PDF (e.g. when $\phi(y)=\delta(y)$) . Moreover, since the slices/projections are one-dimensional the transport map, $f(.,\theta)$, is uniquely defined and the minimization on $f$ has a closed form solution and is calculated from Equation \eqref{eq:1dwp}. The Radon transformations in Equation \eqref{eq:SWM} are: 
\begin{eqnarray}
\left\{\begin{array}{l}
\mathcal{R}I_y(t,\theta)\approx \frac{1}{N}\sum_{n=1}^N \mathcal{R}\phi(t-y_n\cdot\theta,\theta)\\ \\
\mathcal{R}I_x(t,\theta)= \sum_k \frac{\alpha_k}{\sqrt{2\pi\theta^T\Sigma_k\theta}}exp(-\frac{(t-\mu_k\cdot\theta)^2}{2\theta^T\Sigma_k\theta})
\end{array}\right.
\end{eqnarray}

The new formulation avoids the optimization for calculating the Wasserstein distance and enables an efficient implementation for clustering high-dimensional data. Figure \ref{fig:hlidea} demonstrates the high-level idea behind slicing high-dimensional PDFs $I_x$ and $I_y$ and minimizing the p-Wasserstein distance between these slices over GMM components. Moreover, given the high-dimensional nature of the problem estimating density $I_y$ in $\mathbb{R}^d$ requires large number of samples, however,  the projections of $I_y$, $\mathcal{R}I_y(.,\theta)$, are one dimensional and therefore it may not be critical to have large number of samples to estimate these one-dimensional densities. 

\begin{figure*}[t]
\centering
\includegraphics[width=\linewidth]{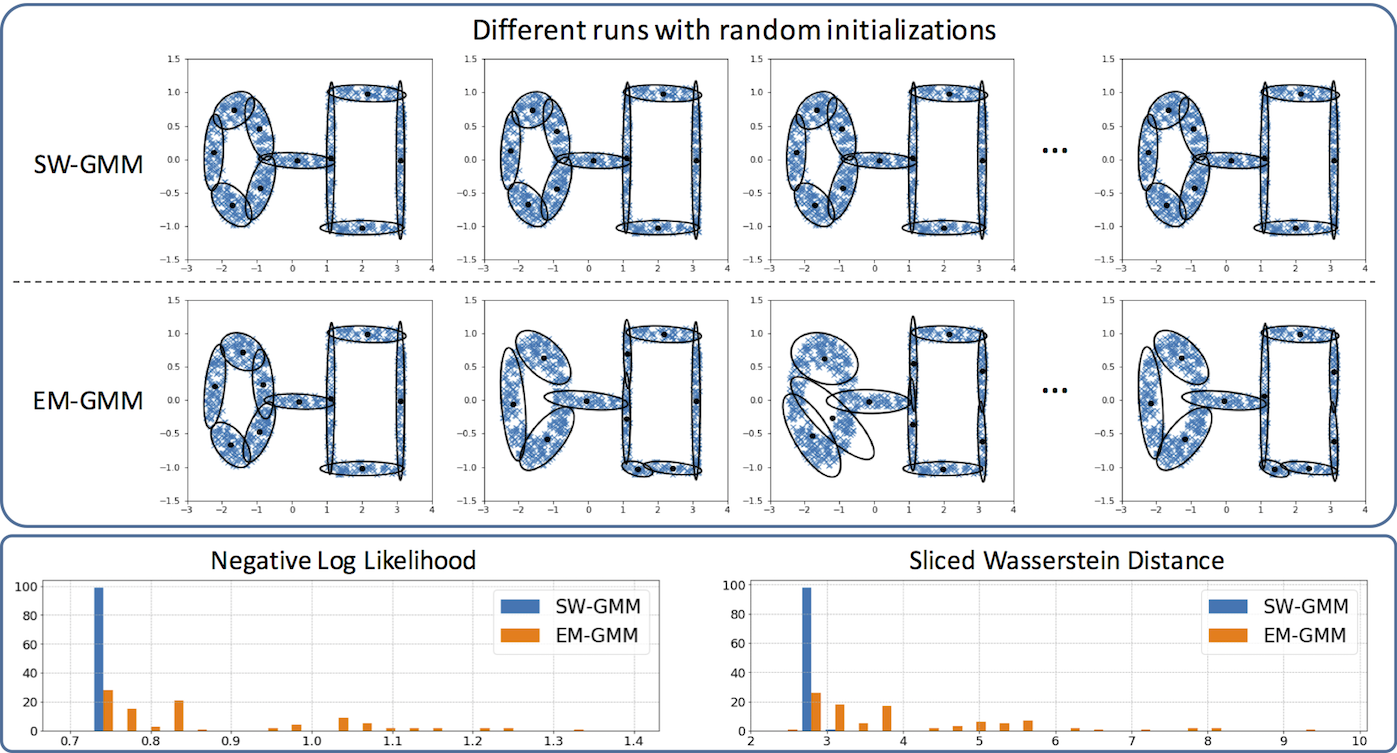}
\caption{Results of 100 runs of EM-GMM and SW-GMM fitting a model with 10 modes to the ring-line-square dataset, showing four samples of random initializations (Top) and histograms across all 100 runs for the negative log-likelihood of the fitted model and the sliced-Wasserstein distance between the fitted model and the data distribution (Bottom). }
\label{fig:exp1}
\end{figure*}

{\bf Optimization scheme:} To obtain a numerical optimization scheme, which minimizes the problem in Equation \eqref{eq:SWM} we first discretize the set of directions/projections. This corresponds to the use of a finite set $\Theta \in \mathbb{S}^{d-1}$, and a minimization of the following energy function, 
\begin{eqnarray}
\operatorname*{inf}_{\mu_k,\Sigma_k,\alpha_k} \frac{1}{|\Theta|}\sum_{l=1}^{|\Theta|}\int_\mathbb{R} |f(t,\theta_l)-t|^p\mathcal{R}I_x(t,\theta_l)dt
\label{eq:SWM_Discrete}
\end{eqnarray}
A fine sampling of $\mathbb{S}^{d-1}$ is required for Equation \eqref{eq:SWM_Discrete} to be a good approximation of \eqref{eq:SWM}. Such sampling, however,  becomes prohibitively expensive for high-dimensional data. Alternatively, following the approach presented in \cite{bonneel2015sliced} we utilize random samples of $\mathbb{S}^{d-1}$ at each minimization step to approximate the Equation \eqref{eq:SWM}. This leads to a stochastic gradient descent scheme where instead of random sampling of the input data, we random sample the projection angles. Finally, the GMM parameters are updated through an EM-like approach where for fixed GMM parameters we calculate the optimal transport map $f$ between random slices of $I_x$ and $I_y$, followed by updating $I_x$ for fixed transport maps $f(.,\theta)$s. Below we describe these steps: 

\begin{enumerate}
\item Generate $L$ random samples from $\mathbb{S}^{(d-1)}$, $\{\theta_1,...,\theta_L\}$.

\item Fix the GMM, $I_x$, and calculate the optimal transport map between slices $\mathcal{R}I_x(\cdot,\theta_l)$ and $\mathcal{R}I_y(\cdot,\theta_l)$ via:
\begin{eqnarray}
f(t,\theta_l)= \mathcal{R}J^{-1}_y(\mathcal{R}J_x(t,\theta_l),\theta_l)
\label{eq:tmap}
\end{eqnarray}
where $\mathcal{R}J_{x(y)}(\cdot,\theta_l)$ is the CDF of $\mathcal{R}I_{x(y)}(\cdot,\theta_l)$. 

\item For fixed transportmaps, $f(\cdot, \theta_l)$s, update the GMM parameters by differentiating Equation \eqref{eq:SWM}:
{\scriptsize\begin{eqnarray*}
&&\frac{\partial SW_p^p}{\partial \alpha_k}= \sum_{l=1}^L \int_\mathbb{R} \frac{|f(t,\theta_l)-t|^p}{\sqrt{2\pi\theta_l^T\Sigma_k\theta_l}}exp(-\frac{(t-\mu_k\cdot\theta_l)^2}{2\theta_l^T\Sigma_k\theta_l}) dt \\
&&\frac{\partial SW_p^p}{\partial \mu_k}= \sum_{l=1}^L\left(\int_\mathbb{R} \frac{\alpha_k|f(t,\theta_l)-t|^p}{\sqrt{2\pi\theta_l^T\Sigma_k\theta_l}}exp(-\frac{(t-\mu_k\cdot\theta_l)^2}{2\theta_l^T\Sigma_k\theta_l})\right.\\ &&\hspace{1.8in}\left.\frac{(t-\mu_k\cdot\theta_l)}{\theta_l^T\Sigma_k\theta_l} dt\right)\theta_l \\
&&\frac{\partial SW_p^p}{\partial \Sigma_k}= \sum_{l=1}^L \left(\int_\mathbb{R} \frac{\alpha_k|f(t,\theta_l)-t|^p}{\sqrt{8\pi(\theta_l^T\Sigma_k\theta_l)^3}}[\frac{(t-\mu_k\cdot\theta_l)^2}{\theta_l^T\Sigma_k\theta_l}-1]\right.\\ &&\hspace{1.2in} \left. exp(-\frac{(t-\mu_k\cdot\theta_l)^2}{2\theta_l^T\Sigma_k\theta_l})dt\right)(\theta_l\theta_l^T)
\label{eq:gradients}
\end{eqnarray*}}
where the summation is over $L$ random projections $\theta_l\in\mathbb{S}^{d-1}$. We use the RMSProp optimizer \cite{tieleman2012lecture}, which provides an adaptive learning rate, to update the parameters of the GMM according to the gradients
\item Project the updated $\Sigma_k$s onto the positive semidefinite cone, and renormalize $\alpha_k$s to satisfy $\sum_k \alpha_k=1$.
\end{enumerate}

Notice that the derivative with respect to the k'th component of the mixture model in Equation \eqref{eq:gradients} is independent of other components. In addition, the transport map for each projection, $f(\cdot,\theta)$, in Equation \eqref{eq:tmap} is calculated independent of the other projections. Therefore the optimization can be heavily parallelized in each iteration. We note that, we have also experimented with the Adam optimizer \cite{kingma2014adam} but did not see any improvements over RMSProp. The detailed update equations are included in the Supplementary materials. In what follows we show the SWM solver for estimating GMM parameters in action.

\begin{figure*}
\includegraphics[width=\linewidth]{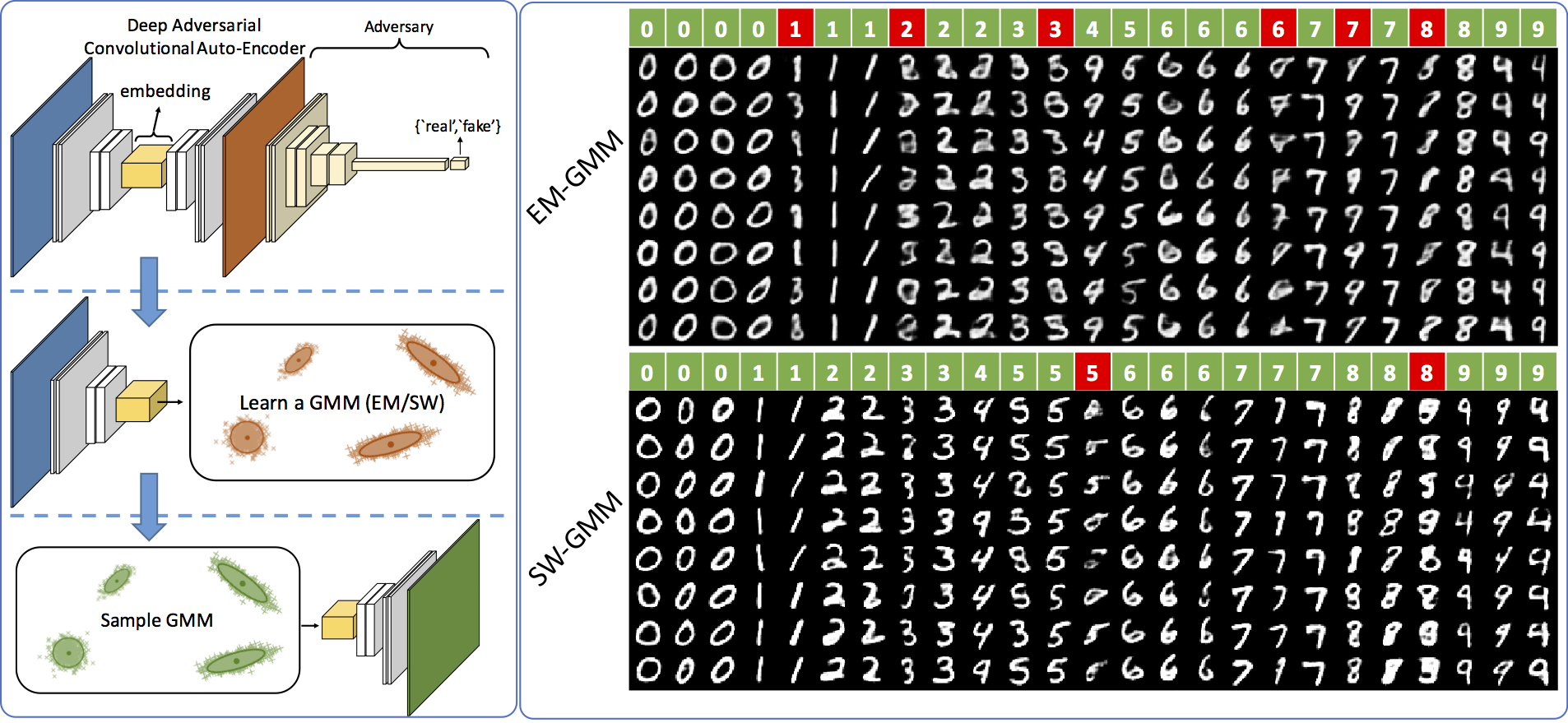}
\caption{Qualitative performance comparison on the MNIST dataset between our method, SW-GMM, and EM-GMM, showing decoded samples for each mode (Right). Modes with bad samples are shown in red. The GMM was applied to a 128-dimensional embedding space (Left).}
\label{fig:mnist}
\end{figure*}

\section{Numerical Experiments}

We ran various experiments on three datasets to test our proposed method for learning GMM parameters. The first dataset is a two-dimensional data-point distribution consisting a ring, a square, and a connecting line (See Figure \ref{fig:exp1}). To show the applicability of our method on higher-dimensional datasets we also used the MNIST dataset \cite{lecun1998mnist} and the CelebFaces Attributes Dataset (CelebA) \cite{liu2015faceattributes}. 

\label{sec:experiments}


\subsection{Robustness to initialization}
We started by running a simple experiment to demonstrate the robustness of our proposed formulation to different initializations. In this test, we used a two-dimensional dataset consisting of a ring, a square, and a line connecting them. For a fixed number of modes, $K=10$ in our experiment, we randomly initialized the GMM. Next, for each initialization, we optimized the GMM parameters using the EM algorithm as well as our proposed method. We repeated this experiment 100 times.

Figure \ref{fig:exp1} shows sample results of the fitted GMM models for both algorithms (Top Row). Moreover, we calculated the histograms of the negative log-likelihood of the fitted GMM and the sliced-Wasserstein distance between the fitted GMM and the empirical data distribution (bottom). It can be seen that our proposed formulation provides a consistent model regardless of the initialization. In $100\%$ of initializations, our method achieved the optimal negative log-likelihood, compared to only $29\%$ for EM-GMM. In addition, both the negative log-likelihood and the sliced-Wasserstein distance for our method are smaller than those of the EM algorithm, indicating that our solution is closer to the global optimum (up to permutations of the modes). 

\subsection{High-dimensional datasets}

We analyzed the performance of our proposed method in modeling high-dimensional data distributions, here, using the MNIST  dataset \cite{lecun1998mnist} and the CelebA dataset \cite{liu2015faceattributes}. To capture the nonlinearity of the image data and boost the applicability of GMMs, we trained an adversarial deep convolutional autoencoder (Figure \ref{fig:mnist}, Left) on the image data. Next, we modeled the distribution of the data in the embedded space via a GMM. The GMM was then used to generate samples in the embedding space, which were consequently decoded to generate synthetic (i.e. 'fake') images. In learning the GMM, we compared the EM algorithm with our proposed method, SW-GMM. We note that the entire pipeline is in an unsupervised learning setting. Figure \ref{fig:mnist} demonstrates the steps of our experiment (Left) and provides a qualitative measure of the generated samples (Right) for the MNIST dataset. It can be seen that the SW-GMM model leads to more visually appealing samples compared to the EM-GMM. In addition, we trained a CNN classifier on the MNIST training data. We then generated 10,000 samples from each GMM component and classified these samples to measure the fidelity/pureness of each component. Ideally, each component should only be assigned to a single digit. We found out that for EM-GMM the components were in average $80.48\%$ pure, compared to $86.98\%$ pureness of SW-GMM components.  

\begin{figure}
\includegraphics[width=\linewidth]{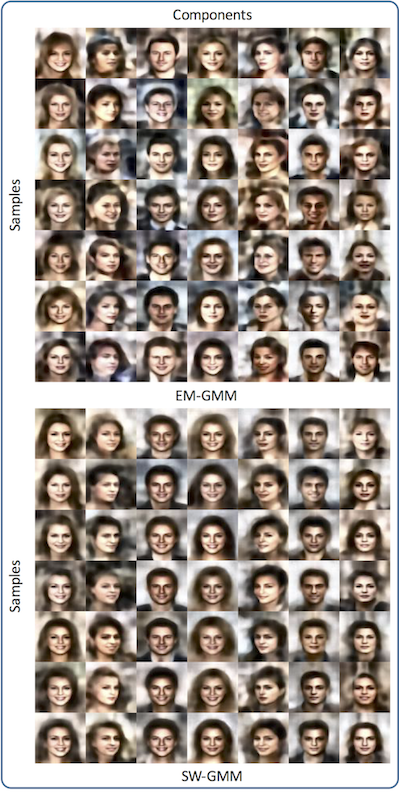}
\caption{Qualitative performance comparison between our method, SW-GMM (Bottom), and EM-GMM (Top), showing decoded samples for several GMM components. The images are contrast enhanced for visualization purposes.}
\label{fig:swemgmm}
\end{figure}

Similarly, a deep convolutional autoencoder was learned for the CelebA face dataset, and a GMM was trained in the embedding space. Figure \ref{fig:swemgmm} shows samples generated from GMM components learned by EM and by our proposed method (The samples generated from all components is attached in the Supplementary materials). We note that, Figures \ref{fig:mnist} and \ref{fig:swemgmm} only provide qualitative measures of how well the GMM is fitting the dataset. Next we provide quantitative measures for the fitness of the GMMs for both methods. 

\begin{figure}
\centering
\begin{minipage}{\columnwidth}
\centering
\includegraphics[width=\columnwidth]{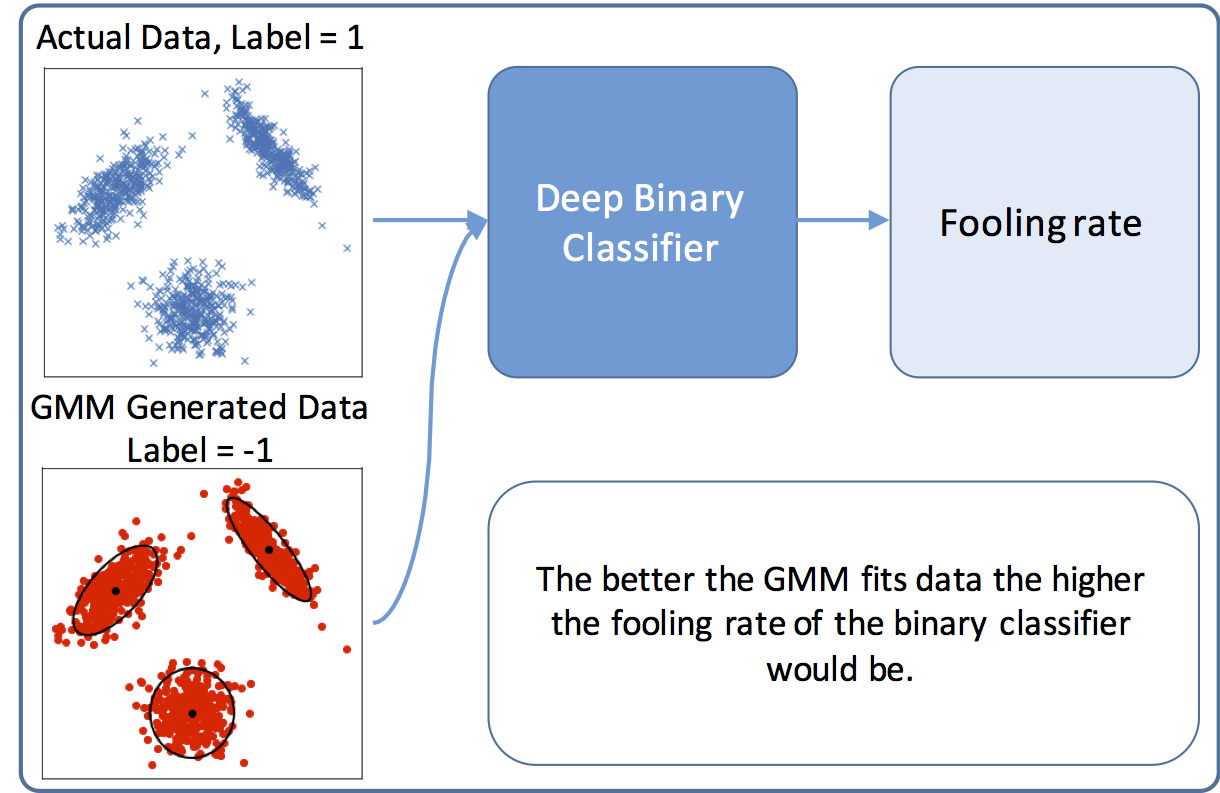}
\end{minipage}
\begin{minipage}{\columnwidth}
\centering
\vspace{.075in}
\renewcommand{\arraystretch}{1.5}
\begin{tabular}{c|c|c|}
\cline{2-3}
 Fooling rate & EM-GMM & SW-GMM \\
\hline
\multicolumn{1}{|c|}{\small Ring-Square-Line}& $46.83\%\pm 1.14\%$& $47.56\%\pm 0.86\%$  \\
\hline
\multicolumn{1}{|c|}{\small MNIST}& $24.87\%\pm 8.39\%$&$41.91\%\pm 2.35\%$  \\
\hline
\multicolumn{1}{|c|}{\small CelebA}& $10.37\%\pm 3.22\%$&$31.83\%\pm 1.24\%$  \\
\hline
\end{tabular}
\end{minipage}
\vspace{.05in}
\caption{A deep discriminator is learned to classify whether an input is sampled from the true distribution of the data or via the GMM. The fooling rate of such a discriminator corresponds to the fitness score of the GMM.}
\label{fig:discriminator}
\end{figure}

We used adversarial training of neural networks \cite{goodfellow2014generative,li2017adversarial} to provide a goodness of fitness of the GMM to the data distribution. In short, we use success in fooling an adversary network as an evaluation metric for goodness of fit of a GMM. A deep discriminator/classifier was trained to distinguish whether a data point was sampled from the actual data distribution or from the GMM. The fooling rate (i.e. error rate) of such a discriminator is a good measure of fitness for the GMM, as a higher error rate translates to a better fit to the distribution of the data. Figure \ref{fig:discriminator} shows the idea behind this experiment, and reports the fooling rates for all three datasets used in our experiments. Note that the SW-GMM consistently provides a higher fooling rate, indicating a better fit to the datasets. Moreover, we point out that while for the low-dimensional ring-square-line dataset both methods provide reasonable models for the dataset, the SW-GMM significantly outperforms EM-GMM for higher-dimensional datasets (i.e. the MNIST and CelebA datasets). 

The details of the architectures used in our experiments are included in the supplementary material.

\section{Discussion}
\label{sec:conclusion}

In this paper, we proposed a novel algorithm for estimating the parameters of a GMM via minimization of the sliced p-Wasserstein distance. In each iteration, our method projects the high-dimensional data distribution into a small set of  one-dimensional distributions utilizing random projections/slices of the Radon transform and estimates the GMM parameters from these one-dimensional projections. While we did not provide a theoretical guarantee that the new method is convex, or that it has fewer local minima, the empirical results suggest that our method is more robust compared to KL-divergence-based methods, including the EM algorithm, for maximizing the log-likelihood function. Consistent with this finding, we showed that the p-Wasserstein metrics result in more well-behaved  energy landscapes. We demonstrated the robustness of our method on three datasets: a two-dimensional ring-square-line distribution and the high-dimensional MNIST and CelebA face datasets. Finally, while we used deep convolutional encoders to provide embeddings for two of the datasets and learned GMMs in these embeddings, we emphasize that our method could be applied to other embeddings including the original data space. 


\section{Acknowledgement}

This work was partially supported by NSF (CCF 1421502). The authors gratefully appreciate countless fruitful conversations with Drs. Charles H. Martin and Dejan Slep\'{c}ev.

{\small
\bibliographystyle{ieee}
\bibliography{SW_GMM.bib}

\begin{thebibliography}{10}\itemsep=-1pt

\bibitem{agueh2011barycenters}
M.~Agueh and G.~Carlier.
\newblock Barycenters in the {W}asserstein space.
\newblock {\em SIAM Journal on Mathematical Analysis}, 43(2):904--924, 2011.

\bibitem{amendola2015maximum}
C.~Am{\'e}ndola, M.~Drton, and B.~Sturmfels.
\newblock Maximum likelihood estimates for gaussian mixtures are
  transcendental.
\newblock In {\em International Conference on Mathematical Aspects of Computer
  and Information Sciences}, pages 579--590. Springer, 2015.

\bibitem{arjovsky2017wasserstein}
M.~Arjovsky, S.~Chintala, and L.~Bottou.
\newblock Wasserstein generative adversarial networks.
\newblock In {\em International Conference on Machine Learning}, pages
  214--223, 2017.

\bibitem{banerjee2005clustering}
A.~Banerjee, S.~Merugu, I.~S. Dhillon, and J.~Ghosh.
\newblock Clustering with bregman divergences.
\newblock {\em Journal of machine learning research}, 6(Oct):1705--1749, 2005.

\bibitem{beecks2011modeling}
C.~Beecks, A.~M. Ivanescu, S.~Kirchhoff, and T.~Seidl.
\newblock Modeling image similarity by gaussian mixture models and the
  signature quadratic form distance.
\newblock In {\em Computer Vision (ICCV), 2011 IEEE International Conference
  On}, pages 1754--1761. IEEE, 2011.

\bibitem{bonneel2015sliced}
N.~Bonneel, J.~Rabin, G.~Peyr{\'e}, and H.~Pfister.
\newblock Sliced and {R}adon {W}asserstein barycenters of measures.
\newblock {\em Journal of Mathematical Imaging and Vision}, 51(1):22--45, 2015.

\bibitem{brenier1991polar}
Y.~Brenier.
\newblock Polar factorization and monotone rearrangement of vector-valued
  functions.
\newblock {\em Communications on pure and applied mathematics}, 44(4):375--417,
  1991.

\bibitem{campbell2006support}
W.~M. Campbell, D.~E. Sturim, and D.~A. Reynolds.
\newblock Support vector machines using gmm supervectors for speaker
  verification.
\newblock {\em IEEE signal processing letters}, 13(5):308--311, 2006.

\bibitem{celik2012automatic}
T.~Celik and T.~Tjahjadi.
\newblock Automatic image equalization and contrast enhancement using gaussian
  mixture modeling.
\newblock {\em IEEE Transactions on Image Processing}, 21(1):145--156, 2012.

\bibitem{chaudhuri2008finding}
K.~Chaudhuri and A.~McGregor.
\newblock Finding metric structure in information theoretic clustering.
\newblock In {\em COLT}, volume~8, page~10, 2008.

\bibitem{chen2017optimal}
Y.~Chen, T.~T. Georgiou, and A.~Tannenbaum.
\newblock Optimal transport for gaussian mixture models.
\newblock {\em arXiv preprint arXiv:1710.07876}, 2017.

\bibitem{chollet2015keras}
F.~Chollet et~al.
\newblock Keras.
\newblock \url{https://github.com/fchollet/keras}, 2015.

\bibitem{cuturi2013sinkhorn}
M.~Cuturi.
\newblock Sinkhorn distances: {L}ightspeed computation of optimal transport.
\newblock In {\em Advances in Neural Information Processing Systems}, pages
  2292--2300, 2013.

\bibitem{cuturi2014fast}
M.~Cuturi and A.~Doucet.
\newblock Fast computation of wasserstein barycenters.
\newblock In {\em International Conference on Machine Learning}, pages
  685--693, 2014.

\bibitem{dempster1977maximum}
A.~P. Dempster, N.~M. Laird, and D.~B. Rubin.
\newblock Maximum likelihood from incomplete data via the em algorithm.
\newblock {\em Journal of the royal statistical society. Series B
  (methodological)}, pages 1--38, 1977.

\bibitem{frogner2015learning}
C.~Frogner, C.~Zhang, H.~Mobahi, M.~Araya, and T.~A. Poggio.
\newblock Learning with a wasserstein loss.
\newblock In {\em Advances in Neural Information Processing Systems}, pages
  2053--2061, 2015.

\bibitem{goldberger2003efficient}
J.~Goldberger, S.~Gordon, and H.~Greenspan.
\newblock An efficient image similarity measure based on approximations of
  kl-divergence between two gaussian mixtures.
\newblock In {\em null}, page 487. IEEE, 2003.

\bibitem{goodfellow2014generative}
I.~Goodfellow, J.~Pouget-Abadie, M.~Mirza, B.~Xu, D.~Warde-Farley, S.~Ozair,
  A.~Courville, and Y.~Bengio.
\newblock Generative adversarial nets.
\newblock In {\em Advances in neural information processing systems}, pages
  2672--2680, 2014.

\bibitem{guerrero2008image}
J.~A. Guerrero-Col{\'o}n, L.~Mancera, and J.~Portilla.
\newblock Image restoration using space-variant gaussian scale mixtures in
  overcomplete pyramids.
\newblock {\em IEEE Transactions on Image Processing}, 17(1):27--41, 2008.

\bibitem{gulrajani2017improved}
I.~Gulrajani, F.~Ahmed, M.~Arjovsky, V.~Dumoulin, and A.~Courville.
\newblock Improved training of wasserstein gans.
\newblock {\em arXiv preprint arXiv:1704.00028}, 2017.

\bibitem{ho2017multilevel}
N.~Ho, X.~Nguyen, M.~Yurochkin, H.~H. Bui, V.~Huynh, and D.~Phung.
\newblock Multilevel clustering via wasserstein means.
\newblock {\em arXiv preprint arXiv:1706.03883}, 2017.

\bibitem{hoffmann05ulva}
H.~Hoffmann.
\newblock {\em Unsupervised Learning of Visuomotor Associations}, volume~11 of
  {\em {MPI} Series in Biological Cybernetics}.
\newblock Logos Verlag Berlin, 2005.

\bibitem{hoffmann05bc}
H.~Hoffmann, W.~Schenck, and R.~M\"{o}ller.
\newblock Learning visuomotor transformations for gaze-control and grasping.
\newblock {\em Biological Cybernetics}, 93:119--130, 2005.

\bibitem{jian2011robust}
B.~Jian and B.~C. Vemuri.
\newblock Robust point set registration using gaussian mixture models.
\newblock {\em IEEE Transactions on Pattern Analysis and Machine Intelligence},
  33(8):1633--1645, 2011.

\bibitem{jin2016local}
C.~Jin, Y.~Zhang, S.~Balakrishnan, M.~J. Wainwright, and M.~I. Jordan.
\newblock Local maxima in the likelihood of gaussian mixture models: Structural
  results and algorithmic consequences.
\newblock In {\em Advances in Neural Information Processing Systems}, pages
  4116--4124, 2016.

\bibitem{kalai2012disentangling}
A.~T. Kalai, A.~Moitra, and G.~Valiant.
\newblock Disentangling gaussians.
\newblock {\em Communications of the ACM}, 55(2):113--120, 2012.

\bibitem{kingma2014adam}
D.~Kingma and J.~Ba.
\newblock Adam: A method for stochastic optimization.
\newblock {\em arXiv preprint arXiv:1412.6980}, 2014.

\bibitem{kolouri2017optimal}
S.~Kolouri, S.~R. Park, M.~Thorpe, D.~Slepcev, and G.~K. Rohde.
\newblock Optimal mass transport: Signal processing and machine-learning
  applications.
\newblock {\em IEEE Signal Processing Magazine}, 34(4):43--59, 2017.

\bibitem{kolouri2015transport}
S.~Kolouri and G.~K. Rohde.
\newblock Transport-based single frame super resolution of very low resolution
  face images.
\newblock In {\em Proceedings of the IEEE Conference on Computer Vision and
  Pattern Recognition}, pages 4876--4884, 2015.

\bibitem{kolouri2016sliced}
S.~Kolouri, Y.~Zou, and G.~K. Rohde.
\newblock Sliced-{W}asserstein kernels for probability distributions.
\newblock In {\em Proceedings of the IEEE Conference on Computer Vision and
  Pattern Recognition}, pages 4876--4884, 2016.

\bibitem{lecun1998mnist}
Y.~LeCun.
\newblock The mnist database of handwritten digits.
\newblock {\em http://yann. lecun. com/exdb/mnist/}.

\bibitem{levy2015}
B.~L{\'e}vy.
\newblock A numerical algorithm for {$L_2$} semi-discrete optimal transport in
  3{D}.
\newblock {\em ESAIM Math. Model. Numer. Anal.}, 49(6):1693--1715, 2015.

\bibitem{li2017adversarial}
J.~Li, W.~Monroe, T.~Shi, A.~Ritter, and D.~Jurafsky.
\newblock Adversarial learning for neural dialogue generation.
\newblock {\em arXiv preprint arXiv:1701.06547}, 2017.

\bibitem{li2013novel}
P.~Li, Q.~Wang, and L.~Zhang.
\newblock A novel earth mover's distance methodology for image matching with
  gaussian mixture models.
\newblock In {\em Proceedings of the IEEE International Conference on Computer
  Vision}, pages 1689--1696, 2013.

\bibitem{liu2015faceattributes}
Z.~Liu, P.~Luo, X.~Wang, and X.~Tang.
\newblock Deep learning face attributes in the wild.
\newblock In {\em Proceedings of International Conference on Computer Vision
  (ICCV)}, 2015.

\bibitem{lucke2017k}
J.~L{\"u}cke and D.~Forster.
\newblock k-means is a variational em approximation of gaussian mixture models.
\newblock {\em arXiv preprint arXiv:1704.04812}, 2017.

\bibitem{mclachlan2014number}
G.~J. McLachlan and S.~Rathnayake.
\newblock On the number of components in a gaussian mixture model.
\newblock {\em Wiley Interdisciplinary Reviews: Data Mining and Knowledge
  Discovery}, 4(5):341--355, 2014.

\bibitem{moeller+04}
R.~M\"oller and H.~Hoffmann.
\newblock An extension of neural gas to local {P}{C}{A}.
\newblock {\em Neurocomputing}, 62:305--326, 2004.

\bibitem{montavon2016wasserstein}
G.~Montavon, K.-R. M{\"u}ller, and M.~Cuturi.
\newblock Wasserstein training of restricted boltzmann machines.
\newblock In {\em Advances in Neural Information Processing Systems}, pages
  3718--3726, 2016.

\bibitem{murphy2012machine}
K.~P. Murphy.
\newblock {\em Machine learning: a probabilistic perspective}.
\newblock MIT press, 2012.

\bibitem{natterer1986mathematics}
F.~Natterer.
\newblock {\em The mathematics of computerized tomography}, volume~32.
\newblock Siam, 1986.

\bibitem{park2017cumulative}
S.~R. Park, S.~Kolouri, S.~Kundu, and G.~K. Rohde.
\newblock The cumulative distribution transform and linear pattern
  classification.
\newblock {\em Applied and Computational Harmonic Analysis}, 2017.

\bibitem{permuter2006study}
H.~Permuter, J.~Francos, and I.~Jermyn.
\newblock A study of gaussian mixture models of color and texture features for
  image classification and segmentation.
\newblock {\em Pattern Recognition}, 39(4):695--706, 2006.

\bibitem{peyre2012wasserstein}
G.~Peyr{\'e}, J.~Fadili, and J.~Rabin.
\newblock Wasserstein active contours.
\newblock In {\em Image Processing (ICIP), 2012 19th IEEE International
  Conference on}, pages 2541--2544. IEEE, 2012.

\bibitem{qian2017deep}
Y.~Qian, E.~Vazquez, and B.~Sengupta.
\newblock Deep geometric retrieval.
\newblock {\em arXiv preprint arXiv:1702.06383}, 2017.

\bibitem{rabin2012wasserstein}
J.~Rabin, G.~Peyr{\'e}, J.~Delon, and M.~Bernot.
\newblock {W}asserstein barycenter and its application to texture mixing.
\newblock In {\em Scale Space and Variational Methods in Computer Vision},
  pages 435--446. Springer, 2012.

\bibitem{rolet2016fast}
A.~Rolet, M.~Cuturi, and G.~Peyr{\'e}.
\newblock Fast dictionary learning with a smoothed wasserstein loss.
\newblock In {\em Artificial Intelligence and Statistics}, pages 630--638,
  2016.

\bibitem{solomon2015convolutional}
J.~Solomon, F.~de~Goes, P.~A. Studios, G.~Peyr{\'e}, M.~Cuturi, A.~Butscher,
  A.~Nguyen, T.~Du, and L.~Guibas.
\newblock Convolutional {W}asserstein distances: Efficient optimal
  transportation on geometric domains.
\newblock {\em ACM Transactions on Graphics (Proc. SIGGRAPH 2015), to appear},
  2015.

\bibitem{thorpe2017transportation}
M.~Thorpe, S.~Park, S.~Kolouri, G.~K. Rohde, and D.~Slep{\v{c}}ev.
\newblock A transportation l\^{} p distance for signal analysis.
\newblock {\em Journal of Mathematical Imaging and Vision}, 59(2):187--210,
  2017.

\bibitem{tieleman2012lecture}
T.~Tieleman and G.~Hinton.
\newblock Lecture 6.5-rmsprop: Divide the gradient by a running average of its
  recent magnitude.
\newblock {\em COURSERA: Neural networks for machine learning}, 4(2):26--31,
  2012.

\bibitem{tipping+99b}
M.~E. Tipping and C.~M. Bishop.
\newblock Mixtures of probabilistic principal component analyzers.
\newblock {\em Neural Computation}, 11:443--482, 1999.

\bibitem{vempala2005random}
S.~S. Vempala.
\newblock {\em The random projection method}, volume~65.
\newblock American Mathematical Soc., 2005.

\bibitem{villani2008optimal}
C.~Villani.
\newblock {\em Optimal transport: old and new}, volume 338.
\newblock Springer Science \& Business Media, 2008.

\bibitem{xiao2017wasserstein}
S.~Xiao, M.~Farajtabar, X.~Ye, J.~Yan, L.~Song, and H.~Zha.
\newblock Wasserstein learning of deep generative point process models.
\newblock {\em arXiv preprint arXiv:1705.08051}, 2017.

\bibitem{yu2012solving}
G.~Yu, G.~Sapiro, and S.~Mallat.
\newblock Solving inverse problems with piecewise linear estimators: From
  gaussian mixture models to structured sparsity.
\newblock {\em IEEE Transactions on Image Processing}, 21(5):2481--2499, 2012.

\end{thebibliography}
}

\clearpage
\section{Supplementary material}
\subsection{Maximum log-likelihood and KL-divergence}
The KL-divergence between $I_x$ and $I_y$ is defined as:
\begin{eqnarray*}
KL(I_x,I_y)=\int_{\mathbb{R}^d} I_y(\rho)log(\frac{I_y(\rho)}{I_x(\rho)})d\rho
\end{eqnarray*}
For the maximum log-likelihood and in the limit and as the number of samples grows to infinity, $N\rightarrow\infty$, we have:
\begin{eqnarray*}
&&\lim_{N\rightarrow\infty}\argmax_{\mu_k,\Sigma_k,\alpha_k} \frac{1}{N}\sum_{n=1}^N log(I_x(y_n))=\\ && \argmax_{\mu_k,\Sigma_k,\alpha_k} \int_{\mathbb{R}^d} I_y(\rho)log(I_x(\rho))d\rho =\\ &&
 \argmin_{\mu_k,\Sigma_k,\alpha_k} -\int_{\mathbb{R}^d} I_y(\rho)log(I_x(\rho))d\rho=\\ &&
\argmin_{\mu_k,\Sigma_k,\alpha_k} \int_{\mathbb{R}^d} I_y(\rho)log(I_y(\rho))d\rho-
\\&& \hspace{1in}
\int_{\mathbb{R}^d} I_y(\rho)log(I_x(\rho))d\rho =\\ &&
\argmin_{\mu_k,\Sigma_k,\alpha_k} \int_{\mathbb{R}^d} I_y(\rho)log(\frac{I_y(\rho)}{I_x(\rho)})d\rho =\\&&
\argmin_{\mu_k,\Sigma_k,\alpha_k} KL(I_x,I_y)
\end{eqnarray*}

\subsection{RMSProp update equations}

SGD often suffers from oscillatory behavior across the slopes of a ravine while only making incremental progress towards the optimal point. Various momentum based methods have been introduced to adaptively change the learning rate of SGD and dampen this oscillatory behavior. In our work, we used RMSProp, introduced by Tieleman and Hinton \cite{tieleman2012lecture}, which is a momentum based technique for SGD. Let $\alpha$ be the learning rate, $\gamma\in(0,1)$ be the decay parameter, and $\kappa$ be the momentum parameter. The update equation for a GMM parameter, here $\mu_k$,  is then calculated from:

\begin{eqnarray}
\left\{\begin{array}{l}
m_k^{(i)}=\gamma m_k^{(i-1)}+(1-\gamma)\frac{\partial SW_p^p(I_x,I_y)}
{\partial \mu_k}\\ \\
g_k^{(i)}=\gamma g_k^{(i-1)}+(1-\gamma)(\frac{\partial SW_p^p(I_x,I_y)}{\partial \mu_k})^2\\ \\
v_k^{(i)}=\kappa v_k^{(i-1)}-\frac{\alpha}{\sqrt{g_k^{(i)}-(m_k^{i})^2+\epsilon}}\frac{\partial SW_p^p(I_x,I_y)}{\partial \mu_k}\\ \\
\mu^{(i)}_k=\mu^{(i-1)}_k+v_k^{(i)}

\end{array}\right.
\end{eqnarray}

Where $m_k$ and $g_k$ are the first and second moments (uncentered variance) of $\partial SW_p^p(I_x,I_y)/\partial \mu_k$, respectively. Similar update equations hold for $\Sigma_k$ and $\alpha_k$.

\subsection{Experimental Details}

\begin{figure*}
\centering
\includegraphics[width=\linewidth]{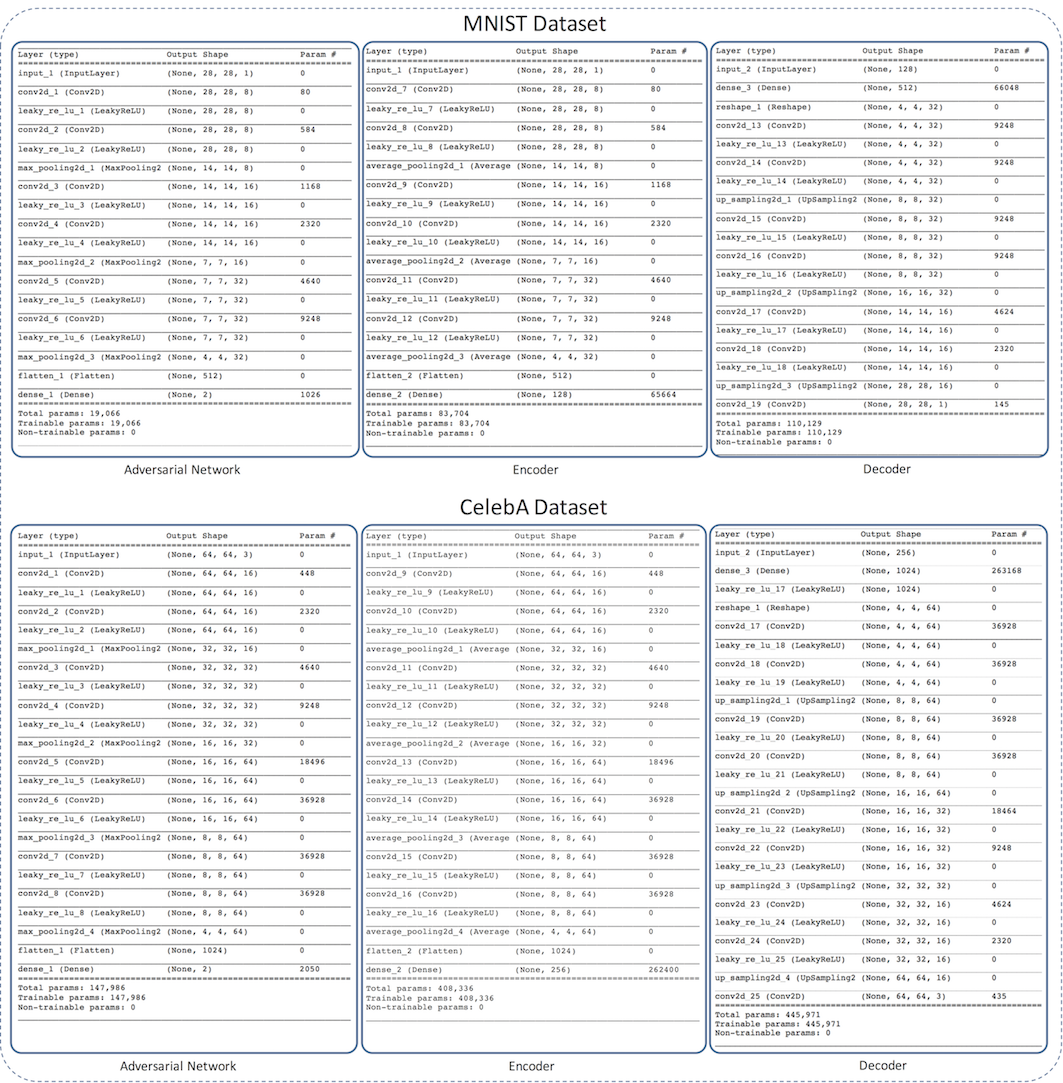}
\caption{Details of the convolutional autoencoders learned for the MNIST and CelebA face dataset}
\label{fig:architectures2}
\end{figure*}

\begin{figure*}
\centering
\includegraphics[width=\linewidth]{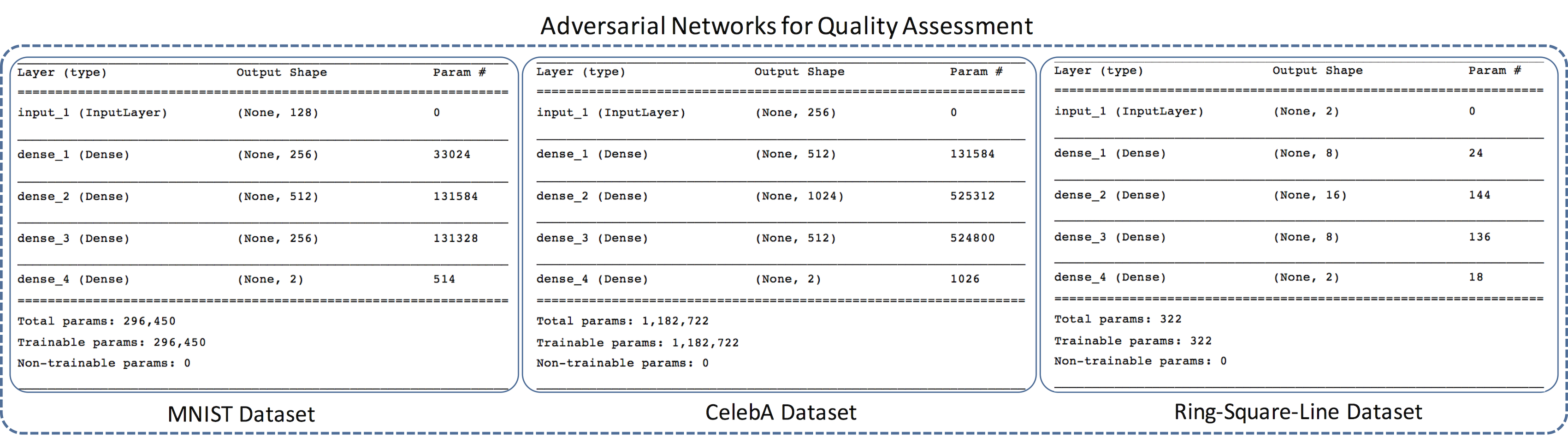}
\caption{Details of the deep binary classifiers used for scoring the fitness of GMMs.}
\label{fig:architectures}
\end{figure*}

Here we provide the detailed implementation and architecture of the adversarial autoencoders we used in our paper. Figure \ref{fig:architectures} shows the detailed architectures of the deep adversarial autoencoder for MNIST and CelebA datasets. The architecture of the deep binary classifiers used for scoring the fitness of the GMMs are shown in Figure \ref{fig:architectures2}. We used Keras \cite{chollet2015keras} for implementation of our experiments. 

For the loss function of the autoencoder we used the {\it mean absolute error} between the input image and the decoded image together with the adversarial loss of the decoded image (equally weighted). The loss functions for training the adversarial networks and the binary classifiers  were chosen to be cross entropy. Finally, we used RMSProp \cite{tieleman2012lecture} as the default optimizer for all the models, and trained the models over $100$ Epochs, with batch size of $250$. 

\subsection{CelebA Generated Images}

\begin{figure*}
\centering
\begin{minipage}{\linewidth}
\centering
\includegraphics[width=\columnwidth]{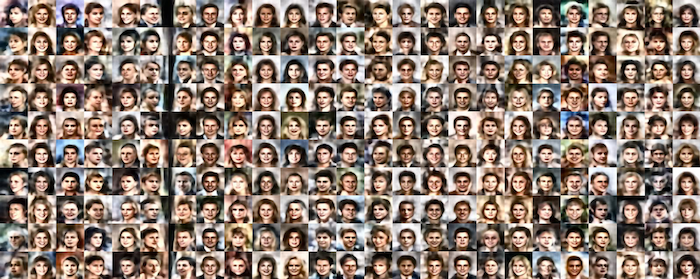}
(a)
\end{minipage}
\\
\begin{minipage}{\linewidth}
\centering
\includegraphics[width=\columnwidth]{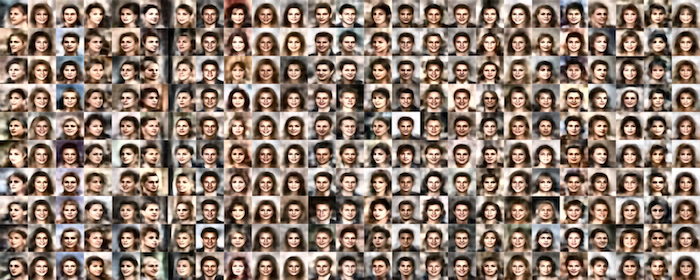}
(b)
\end{minipage}
\caption{GMM Samples Generated from the GMM learned from EM-GMM (a), and from SW-GMM (b). Each column depicts random samples from a single component of the GMM.}
\label{fig:fullresults}
\end{figure*}

Figure \ref{fig:fullresults} shows all the GMM components learned by EM and our SWM formulation. 

\end{document}